\documentclass[10pt, a4paper]{article}

\usepackage[final]{lrec2026} 

\usepackage{cleveref}
\usepackage{enumitem}

\title{Transparency as Architecture: Structural Compliance Gaps in \linebreak EU AI Act Article~50~II}

\name{Vera Schmitt\textsuperscript{$\star$,$1$, $2$, $3$}\thanks{\hspace{-.5cm}\textsuperscript{$\star$}Both authors contributed equally to this research.}, Niklas Kruse\textsuperscript{$\star$,$4$}, Premtim Sahitaj\textsuperscript{$1$}, Julius Schöning\textsuperscript{$4$}} 

\address{Technichal University Berlin XpaliNLP Group\textsuperscript{$1$}, Research Center for Artificial Intelligence (DFKI) \\ Berlin\textsuperscript{$2$}, Center for European Research on Trusted AI (CERTAIN)\textsuperscript{$3$}, Faculty of \\ Engineering and Computer Science, Osnabrück University of Applied Sciences\textsuperscript{$4$} \\
         \{vera.schmitt, premtim.sahitaj\}@tu-berlin.de,  \{niklas.kruse,j.schoening\}@hs-osnabrueck.de\\}

\abstract{
Art. 50 II of the EU Artificial Intelligence Act mandates dual transparency for AI-generated content: outputs must be labeled in both human-understandable and machine-readable form for automated verification. This requirement, entering into force in August 2026, collides with fundamental constraints of current generative AI systems. Using synthetic data generation and automated fact-checking as diagnostic use cases, we show that compliance cannot be reduced to post-hoc labeling. In fact-checking pipelines, provenance tracking is not feasible under iterative editorial workflows and non-deterministic LLM outputs; moreover, the assistive-function exemption does not apply, as such systems actively assign truth values rather than supporting editorial presentation. In synthetic data generation, persistent dual-mode marking is paradoxical: watermarks surviving human inspection risk being learned as spurious features during training, while marks suited for machine verification are fragile under standard data processing. Across both domains, three structural gaps obstruct compliance: (a) absent cross-platform marking formats for interleaved human-AI outputs; (b) misalignment between the regulation's 'reliability' criterion and probabilistic model behavior; and (c) missing guidance for adapting disclosures to heterogeneous user expertise. Closing these gaps requires transparency to be treated as an architectural design requirement, demanding interdisciplinary research across legal semantics, AI engineering, and human-centered design.
 \\ \newline \Keywords{AI Act, transparency, legal compliance, data governance, synthetic data, fact-checking} }

\begin{document}

\maketitleabstract

\section{Introduction}
The proliferation of generative AI systems across high-stakes domains such as journalism, scientific research, public administration, and automated decision-making has made the provenance and authenticity of digital content a pressing governance concern. Synthetic text, images, and structured data can now be produced at scale, at low cost, and with a degree of realism that makes human detection increasingly unreliable~\cite{schmitt2025TransparencyEvaluatingExplainability,Varanasi2023}. In response, the European Union has enacted the Artificial Intelligence Act~(AI Act), which represents a decisive shift from \textit{ex post} regulation of digital technologies toward the \textit{ex ante} governance of AI systems~\cite{Kruse2025}. Unlike earlier frameworks such as the General Data Protection Regulation~(GDPR), which targets data processing and individual rights, the AI Act directly regulates the behavior, outputs, and societal effects of AI systems, most directly through Art.~50~II, which enters into force in August 2026. 

Art.~50~II imposes a dual transparency requirement on providers of generative AI: outputs must be labeled in a human-understandable manner \textit{and} in a machine-readable form that enables automated verification. The provision applies across data types and deployment contexts, from text and images to synthetic datasets used in AI training pipelines. Its regulatory intent is clear: restoring epistemic trust in digital content by making AI involvement permanently traceable. Its technical specifications, however, are absent. Art.~50~II mandates that labeling solutions be effective, interoperable, robust, and reliable, while referring to watermarks, metadata labels, and cryptographic methods as candidate approaches. None of these criteria are defined operationally, and the technical standards that would supply such definitions are still under development~\cite{ec2025codeofpractice}. 

This paper argues that operationalizing Art.~50~II cannot be achieved through labeling solutions appended to existing systems. Generative AI operates under constraints that are structurally at odds with the regulation's requirements. Probabilistic outputs resist deterministic attribution; provenance chains fragment when human and AI contributions are interleaved; and the regulation's demand for reliable and effective transparency provides no guidance on what these properties mean for systems whose outputs vary across identical inputs. These are not implementation difficulties to be engineered away. They reflect conceptual mismatches between legal specification and technical reality that require coordinated research across disciplines. 

To expose and structure these mismatches, we analyze two technically distinct but jointly revealing use cases. In \textit{synthetic data generation}, dual-mode marking introduces a paradox between labeling persistence and training data integrity: watermarks designed to survive human inspection risk being learned as spurious features during model training, while marks suited for machine verification are fragile under standard data processing operations. In \textit{automated fact-checking}, RAG-based multi-source attribution cannot be adequately represented in existing metadata schemas such as Dublin Core or Schema.org, and iterative editorial workflows degrade marking signals to the point where human and AI contributions can no longer be cleanly separated. Beyond these technical limitations, the Art.~50~II assistive-function exemption does not apply to fact-checking systems, as they actively assign truth values and confidence judgments to claims rather than merely supporting editorial presentation, a legal distinction with direct compliance consequences. Together, these cases expose three structural gaps that any compliance pathway must address. 

To systematically address these tensions, two research questions are proposed that serve as the foundation for our analysis.
\begin{itemize}[itemindent=0.5cm] 
    \item[\textbf{RQ\,1:}] Which requirements of Art.~50~II (human-understandability, machine-readability, effectiveness, interoperability, robustness, and reliability) can be technically implemented within current AI systems, and where do systematic, non-incidental limitations arise? 
    \item[\textbf{RQ\,2:}] How do legal concepts such as `understandability' and `reliability' diverge from their technical counterparts, explainability, quality assurance, and output consistency, and what are the compliance consequences of these divergences for practitioners and regulators? 
\end{itemize} 

\Cref{sec:50II} situates Art.~50~II within the AI Act's regulatory architecture and examines its technical requirements. \Cref{sec:UC} examines both research questions through the lens of two high-stakes use cases, assessing where compliance is feasible and where it breaks down. \Cref{sec:Pathways} synthesizes the three structural gaps and identifies preliminary compliance pathways grounded in the use case analysis. \Cref{sec:conclu} concludes by repositioning transparency from a post-hoc metadata problem to an architectural design requirement, and outlines what a research agenda would need to deliver before Art.~50~II takes effect.

\section{Art.~50~II as Technical Requirement}\label{sec:50II}
Art.~50 of the AI Act establishes transparency obligations for providers and deployers of certain AI systems, sitting within Chapter~IV and entering into force on 2~August~2026. The provision addresses four distinct scenarios: AI systems that interact directly with natural persons (Art.~50~I); systems that generate synthetic audio, image, video or text content (Art.~50~II); emotion recognition and biometric categorisation systems (Art.~50~III); and systems that generate or manipulate deep fake content or text published for public information purposes (Art.~50~IV). This paper focuses on Art.~50~II, which imposes the most technically demanding obligations and applies directly to providers of generative AI systems, including general-purpose AI models. 

Art.~50~II requires providers of systems generating synthetic audio, image, video or text content to ensure that outputs are marked in a machine-readable format and detectable as artificially generated or manipulated. Crucially, the obligation is qualified: providers must ensure their technical solutions are effective, interoperable, robust and reliable \textit{as far as this is technically feasible}, taking into account the specificities and limitations of various types of content, the costs of implementation, and the generally acknowledged state of the art as reflected in relevant technical standards. Three exemptions limit the obligation's scope. First, it does not apply where AI systems perform an assistive function for standard editing. Second, it does not apply where the system does not substantially alter the input data or its semantics. Third, it does not apply where use is authorised by law for purposes of detecting, preventing, investigating or prosecuting criminal offences. The interaction between these exemptions and the use cases examined in this paper is non-trivial and is addressed in Section~\ref{sec:UC}. 

The regulation is formulated in a deliberately technology-neutral manner: Art.~50~II names no specific technical implementation and provides no operational definition of the four quality criteria. Recital~133 identifies candidate approaches, including watermarks, metadata labels, and cryptographic methods, but does not specify formats, protocols or measurement criteria. This neutrality creates a structural compliance problem. A provider cannot determine from the regulation alone whether a given watermarking scheme satisfies the effectiveness criterion, what interoperability requires across platforms and jurisdictions, how robustness should be measured under realistic processing conditions, or what reliability means for systems whose outputs are probabilistic and non-deterministic. These are not peripheral questions: they determine whether any specific technical implementation is compliant. 

The regulation's response to this uncertainty is to defer to technical standards, referenced in Art.~50~II as the intended source of operational guidance. Art.~50~VII further provides that the AI Office shall facilitate codes of practice to support effective implementation, and that the Commission may adopt implementing acts to approve or specify common rules for these obligations. However, the relevant harmonized standards remain under development by international, European and national standardization bodies, and it is unclear whether they will be finalized before the provision enters into force~\cite{ec2025codeofpractice}. The standards available to date address adjacent concerns rather than output-level transparency. ISO~42001 governs organizational AI management systems. ISO/IEC~24028 addresses trustworthiness at the system level. ISO/IEC~24027 provides methodologies for bias assessment. None specifies how a label should be designed, embedded, or verified for a given class of AI-generated output, nor how persistence should be maintained once an output passes through downstream processing, editing or format conversion. 

This gap between regulatory requirement and available technical guidance is structural rather than incidental. The dual transparency obligation, which this paper reads as requiring both a human-understandable and a machine-readable marking, applies across a heterogeneous class of data types, including text, images, audio and structured datasets, each of which presents distinct technical constraints. The absence of harmonized, output-level standards leaves providers without a clear compliance pathway, and the technically feasible qualification in Art.~50~II, while pragmatic, introduces its own ambiguity: it is unclear who determines feasibility, by what standard, and at what point in the system life-cycle. These unresolved questions set the stage for the specific technical breakdowns examined in the use cases that follow.

\section{Use Cases Influenced by Art.~50~II}\label{sec:UC}
This section examines the operational challenges of Art.~50~II through two high-stakes use cases: synthetic data generation and automated fact-checking. These cases are technically distinct but jointly detecting similar issues. Each exposes a different facet of the compliance gap identified in Section~\ref{sec:50II}: the synthetic data case reveals how the dual transparency obligation conflicts with the technical requirements of model development pipelines, while the fact-checking case reveals how iterative human-AI workflows undermine provenance tracking and why the assistive-function exemption does not provide support. Together, they ground the three structural gaps described further in Section~\ref{sec:Pathways}.

\begin{figure*}[!ht]
\begin{center}
\includegraphics[width=.9\linewidth]{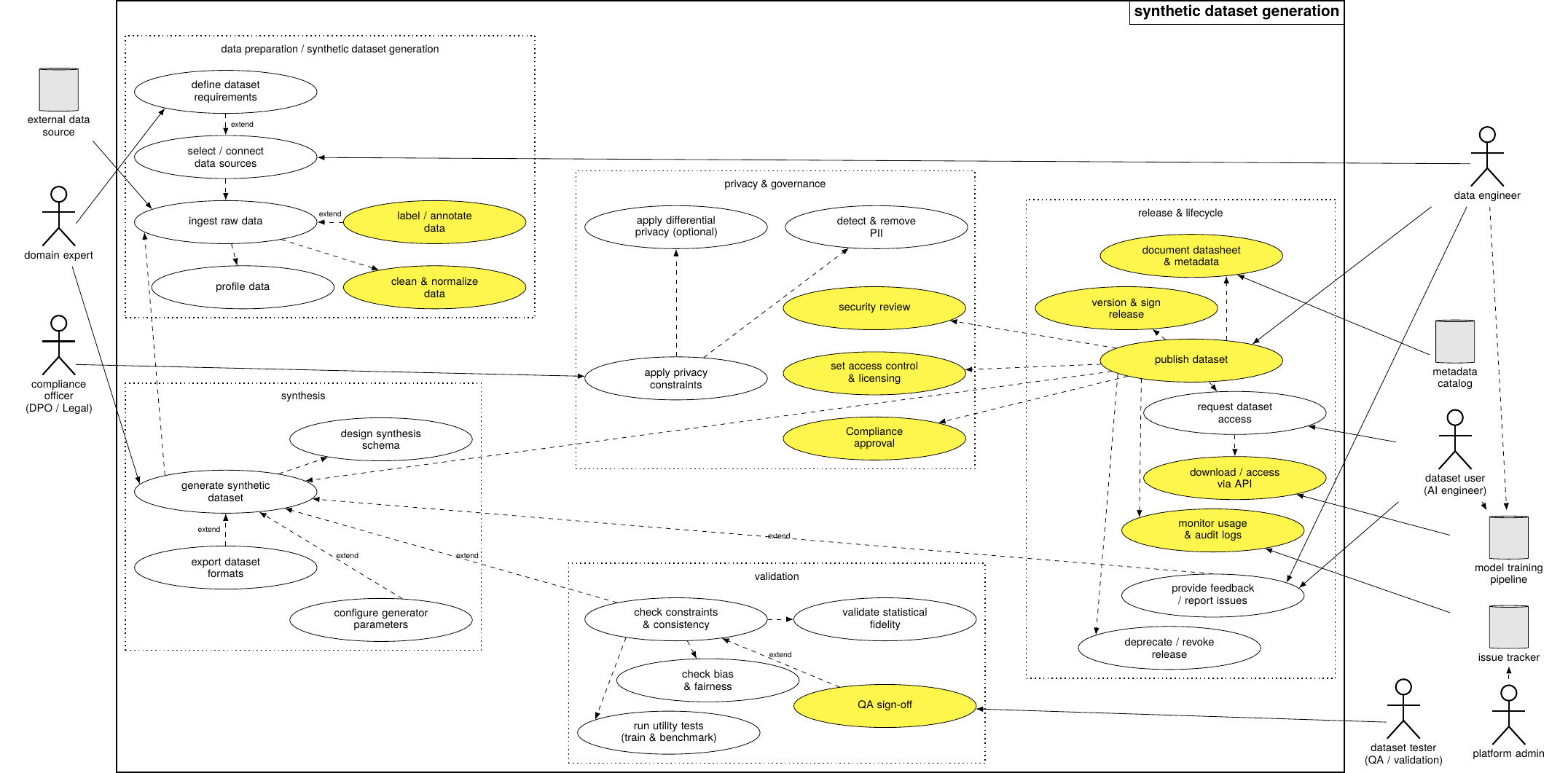}
\caption{Use case diagram of a synthetic data generation system; the yellow-highlighted use case might be relevant for fulfilling Art.~50~II. Note that, depending on the system design, some use cases may be more or less relevant to Art.~50~II.}
\label{fig:UC1}
\end{center}
\end{figure*}

\subsection{Synthetic Data Generation Systems}\label{sec:SyntheticDataGeneration}

The availability of data for developing high-performance AI systems remains a significant challenge across both academic and economic fields~\cite{Li2021,Sambasivan2021,Malerba2024}. 
Modern AI systems, particularly those employing transformer architectures~\cite{Kaplan2020,Richter2022,Halevy2009}, 
necessitate substantial datasets to attain optimal performance. Conventionally, data has been obtained from the real world, a process frequently associated with high costs and limited availability~\cite{Buhrmester2011}. These facts suggest that the development of next-generation AI systems is encountering a significant impediment: insufficient data to enhance their capabilities. A potential solution to this challenge is to augment existing datasets with synthetic data~\cite{Mumuni2022,Wachter2025}.

As illustrated in \Cref{fig:UC1}, the generation of synthetic data encompasses various stakeholders and a range of digital inputs and outputs, including one or several datasets, AI modeling interfaces, and licenses. The synthetic data itself can manifest in various forms, including texts, images, and time series. Although disparate data types are derived from reality, they do not exist outside the synthetic data space. Synthetic data offers a variety of added values. Real-world data can be collected through direct or indirect methods, both of which are derived from reality. These datasets are often subject to real-world biases and can be expensive or time-consuming~\cite{Wachter2024,Gebru2021}. Consequently, existing feature gaps or biases can be addressed by leveraging synthetic data. Another potential application is the pre-training of large AI models, in which synthetic features are used to train fundamental features, thereby improving performance on real datasets. A salient question that remains unanswered is the extent to which synthetic data can completely replace real-world data. In the context of human subjects research, synthetic data often exhibits a high degree of indistinguishability from artificial data, which falls under Art.~50~II if generative AI is used to create it. However, given the inherent variability of image data, particularly in its presentation and design, it is important to note that, e.g., synthetic image data often contains features that are more readily identifiable by an AI system than by humans. Consequently, a discrepancy arises between synthetic and real data, known as the ``reality gap'' or ``domain gap'' . \cite{Wachter2025,Peng2018}. The extent of this discrepancy remains a subject of ongoing research and will be important in determining whether synthetic data needs to be labeled in the context of Art.~50~II.

\subsubsection{Art.~50~II Analysis and Applicability} \label{sec:ucArt50}
A system for the generation of synthetic data, as illustrated in \Cref{fig:UC1}, constitutes a standard case of Art.~50~II. Such systems overlap with the regulatory aspects of Art.~50~IV, which refers to deepfakes. Whether an AI system is subject to Art.~50~II or IV depends largely on the type of data it generates. According to Recital~134, a deepfake is characterized by the fact that the AI output is intended to falsely convince the recipient that the altered output is actual reality. If synthetic generation is used merely to substitute for the dataset creation process and identical artificial representations of real objects are created, this is likely to constitute a case under Art.~50~IV. This generation process could constitute a deepfake within the meaning of Recital 134. If, instead, data is generated to fill gaps in a dataset that cannot be collected because such data does not exist in a usable form, or to expand an existing dataset, then no deepfake is created; rather, new content is created. Such synthetic data generation systems are regulated under Art.~50~II. As depicted in \Cref{fig:UC1}, the use cases within the subsystem data synthesis do not fall under Art.~50~II. However, within the subsystem synthetic dataset generation, three use cases ``label \& annotate data'' clean \& normalize'' , and ``define dataset requirements'' ,  constitute a standard case of Art.~50~II, since the dataset output must be labeled in a way that is both machine-readable and accessible to humans, e.g., using the watermarks mentioned in Recital~133. However, if synthetic datasets are used to train AI systems, the machine-readability of watermarks can become an obstacle to the usability of synthetic data for these purposes. Since in these cases the watermark occurs very frequently, depending on the proportion of synthetic data in the entire dataset, there could be a risk that, due to the distribution of the feature, the AI system treats the watermark as a feature relevant to the training process rather than the actual training content. This links to the subsystem ``release \& life-cycle,'' in which several use cases need to comply with Art.~50~II, e.g., to offer two versions of such synthetic data via an application programming interface (API) so that watermarks visible to an AI can be removed before training, enabling an unbiased training of the AI architecture. The version intended for human recipients contains a human-recognizable watermark instead~\cite{Simmons2024,Bohacek2025,Kruse2024}. Such watermark methods allow the standard's purpose to continue to be fulfilled while the whole life-cycle and privacy governance ensure that synthetic data generation systems are usable and compliant with Art.~50~II.

\subsubsection{Technical and Operational Challenges}
Implementing Art.~50~II's dual transparency requirements for synthetic data generation systems presents technical and operational challenges within the subsystems ``release \& life-cycle'' , ``synthetic dataset generation'' , and privacy \& governance'' . While watermarking techniques, as referenced in Recital~133, appear to offer a straightforward technical solution with permanently linked labels to the data~\cite{Militsyna2025,Labuz2024}, their practical implementation reveals significant complexities and will hinder the use of synthetic datasets. 

Firstly, the fundamental tension between human-readable labeling and machine-readability engenders a technical paradox~\cite{Militsyna2025}. Watermarks designed to be discernible to humans, e.g., subtle visual patterns or embedded text, often impede training when synthetic data is used to train AI models. As discussed in \Cref{sec:ucArt50}, watermarks have the potential to evolve into features that AI systems learn to recognize rather than disregard, thereby jeopardizing the integrity of the training data. Jeopardizing the integrity of the training data creates a critical conflict: the regulatory requirement for persistent labeling directly contradicts the technical need for clean, unadulterated data in model training pipelines~\cite{Wachter2025,Chen2024,Geirhos2020}.
The proposed solution of maintaining two versions of synthetic datasets for the subsystem's ``release \& life-cycle'' use cases introduces operational complexity, requiring additional data management infrastructure and potentially increasing data providers' costs.

Secondly, contemporary watermarking technologies are deficient in terms of the robustness required for Art.~50~II's ``reliable'' standard. Watermarks are often susceptible to compromise when, e.g., subjected to conventional image processing operations such as compression, resizing, cropping, and format conversion. These operations are frequently employed in data pipelines~\cite{Chen2024,Wan2022}. This fragility undermines the ``persistence'' requirement of Art.~50~II, as watermarks may be inadvertently removed or altered during standard data processing. Furthermore, the efficacy of watermarking varies considerably across different data types and content~\cite{Fernandez2023}, with complex or highly unbalanced data posing particular challenges for reliable watermark embedding and detection.

Thirdly, the absence of interoperable technical standards engenders a fragmented implementation landscape~\cite{Simmons2024,Bohacek2025}. Although Recital~133 references watermarks, metadata labels, and cryptographic methods, it does not provide detailed specifications regarding technical formats or protocols. This will result in a proliferation of proprietary watermarking solutions that lack cross-platform compatibility. This is why the use cases ``label \& annotate data'' clean \& normalize'' , and ``define dataset requirements'' also need to consider Art.~50~II. To illustrate, a watermark embedded using one provider's technology may not be detectable by another provider's verification system, thereby violating the ``interoperable'' requirement of Art.~50. This fragmentation engenders substantial impediments for organizations that must integrate synthetic data from multiple sources or utilize data across disparate AI systems.

Fourthly, the scope ambiguity of Art.~50~II exacerbates these technical challenges. The regulatory framework lacks clarity on the necessity of labeling synthetic data used for AI training in a manner consistent with synthetic data distributed to end users. This ambiguity creates operational uncertainty for the system compliance office, which must decide when a watermark is human-recognizable under AI Act Art.~50 and the Accessibility Act. The absence of regulatory guidance on this distinction compels organizations to make potentially costly implementation decisions without clear legal direction.

The tension between regulatory requirements and practical AI system development presents a significant operational challenge. Data scientists and AI engineers generally prioritize data quality and model performance over compliance considerations~\cite{Rakova2021,Varanasi2023,Sambasivan2021}. The implementation and maintenance of dual labeling systems within the ``release \& life-cycle'' subsystem may be a bottleneck to innovation rather than a necessary compliance measure.

The aforementioned challenges collectively demonstrate that Art.~50~II's transparency requirements cannot be met by simple technical add-ons to existing data-generation pipelines. Instead, a fundamental rethinking of the generation, labeling, and management of synthetic data throughout its life-cycle is necessary.

\begin{figure*}[!ht]
\begin{center}
\includegraphics[width=.9\linewidth]{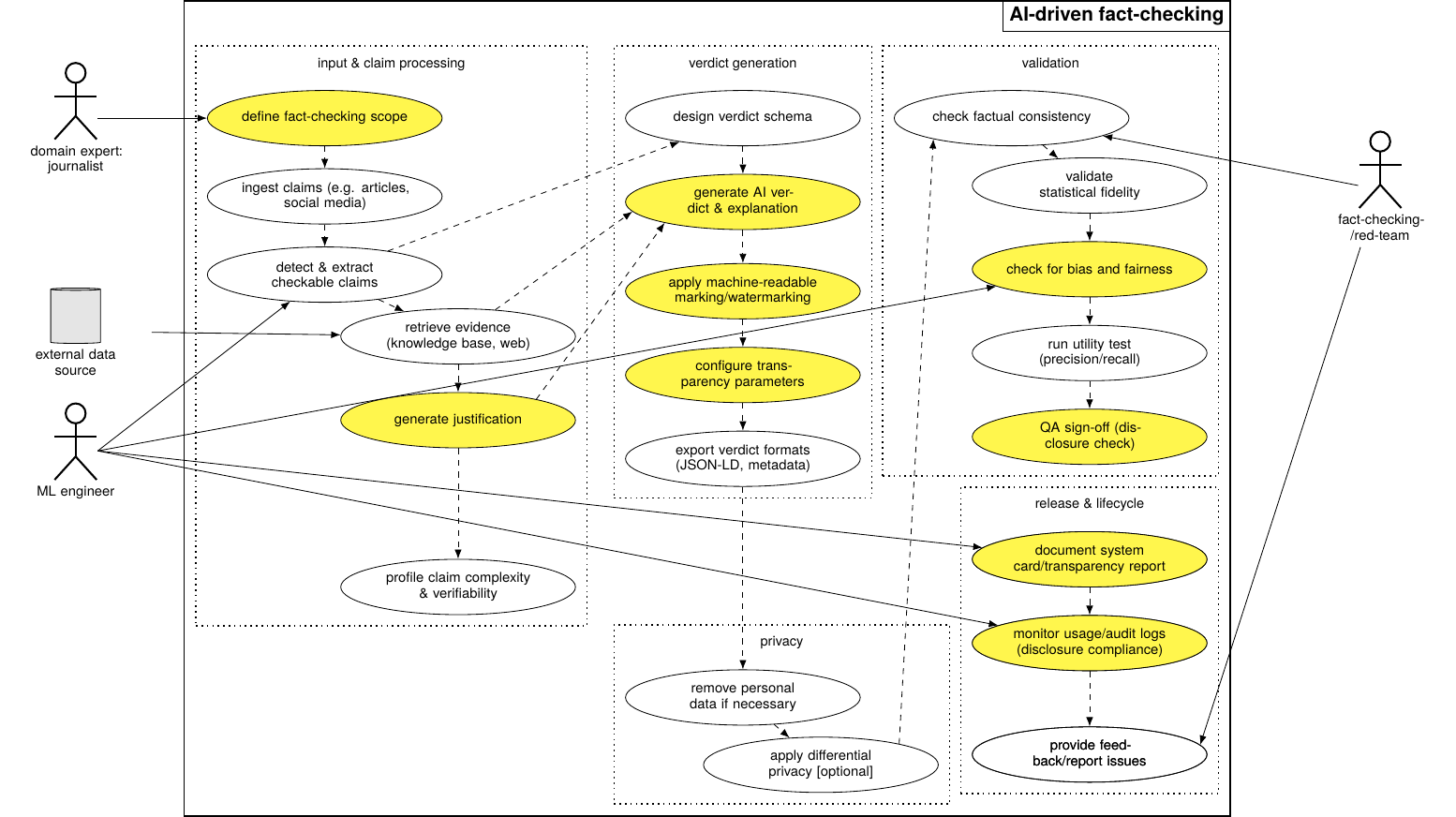}
\caption{Use case diagram of an AI-driven fact-checking system; the yellow-highlighted use case might be relevant for fulfilling Art.~50~II. Note that, depending on the system design, some use cases may be more or less relevant to Art.~50~II.}
\label{fig:UCfacts}
\end{center}
\end{figure*}

\subsection{Fact-Checking Systems}
Fact-checking systems constitute a paradigmatic application domain in which the transparency obligations of Art.~50~II intersect with highly complex technical and editorial workflows. Contemporary automated systems~\cite{guo2022SurveyAutomatedFactChecking} increasingly rely on large language models (LLMs) embedded in multi-stage pipelines that typically comprise claim detection~\cite{hassan2017claimbuster}, evidence retrieval~\cite{sahitaj2025AutomatedFactCheckingRealWorlda}, veracity assessment, and the retrieval-augmented generation (RAG)~\cite{lewis2020retrieval} of textual justifications or summaries. These pipelines connect content platforms, the actors who use the system's outputs, like journalists or moderators, and the organizations that deploy it, such as newsrooms, turning verified claims into reports that can determine moderation or editorial decisions~\cite{schlichtkrull2023IntendedUsesAutomated}. When LLMs are used in automated fact-checking pipelines, it is often difficult to understand why the system reached a specific verdict. Explanations can end up sounding like after-the-fact analyses rather than justifying the actual reasoning behind claim-verification decisions~\cite{tan2025ImprovingExplainableFactChecking}. User studies also suggest that explanations may improve perceived clarity without reliably improving AI-assisted human decision quality, and can even increase overconfidence~\cite{schmitt2025TransparencyEvaluatingExplainability}, indicating that transparency mechanisms can actively undermine their own regulatory purpose if poorly designed

\subsubsection{Art.~50~II Analysis and Applicability}
As illustrated in Fig.~\ref{fig:UCfacts}, fact-checking systems fall within the scope of Art.~50~II whenever they generate textual justifications for veracity assessments, synthesize evidence from multiple sources, or disseminate AI-generated evaluations on matters of public interest. The boundary between Art.~50~II and the assistive-function exemption in Art.~50~II sentence~3 turns on semantic transformation. The exemption covers systems that perform standard editing functions or do not substantially alter the semantics of their inputs. Fact-checking systems do neither: they actively transform user-submitted claims by assigning truth values, confidence scores, and evidential justifications. This constitutes a substantive semantic transformation rather than editorial assistance and therefore falls outside the exemption's scope. Unlike the synthetic data case, where the Art.~50~II versus Art.~50~IV boundary depends on intent and realism, the fact-checking case involves a different boundary question: whether the assistive-function exemption applies. The answer is no, and this has direct compliance consequences: providers cannot reduce their marking obligations through the exemption and must implement full dual transparency for all pipeline outputs. 

Within the use case architecture depicted in Fig.~\ref{fig:UCfacts}, this obligation maps onto specific subsystems. In the verdict generation subsystem, the use cases \textit{generate AI verdict and explanation} and \textit{apply machine-readable marking and watermarking} constitute the primary compliance locus: outputs must be marked in a machine-readable format and detectable as artificially generated at the point of generation. In the release and lifecycle subsystem, the use cases \textit{monitor usage and audit logs} and \textit{document system card and transparency report} carry secondary compliance obligations, as they must preserve evidence of AI involvement across the output lifecycle. However, a critical complication arises from the interleaving of human and AI contributions. When a journalist reviews, edits, and publishes an AI-generated verdict, the boundary between AI-generated and human-authored content becomes impossible to demarcate cleanly. Unlike the synthetic data case, where a dual-version API can separate watermarked and watermark-free outputs for different recipients, no equivalent architectural solution exists for fact-checking pipelines: the human editorial process itself destroys the provenance chain that machine-readable marking requires. 

Current governance standards exhibit the same structural gap identified in Section~\ref{sec:50II} and Section~\ref{sec:SyntheticDataGeneration}. ISO/IEC~42001:2023 addresses organizational AI management rather than output-level transparency. ISO/IEC~24027:2021 provides bias assessment methodologies relevant to source selection and ideological bias in fact-checking, but does not mandate user-facing disclosure of those biases. IEEE~P7001 proposes graded transparency levels that remain misaligned with Art.~50~II's simultaneous human-readable and machine-readable marking requirement. The question of risk classification adds a further compliance dimension. The societal impact of automated fact-checking, particularly in electoral contexts, suggests potential qualification as a high-risk system under Annex~III, Art.~8~aa~\cite{schmitt2024implications}, which would trigger full Chapter~III obligations including conformity assessment, human oversight under Art.~14, and technical documentation under Art.~11. Journalist-facing tools functioning in an advisory rather than decisive capacity may alternatively qualify as limited-risk assistive systems, in which case compliance is largely confined to Art.~50's transparency requirements. This ambiguity is not merely theoretical: the applicable compliance pathway and associated architectural burden differ substantially between these two classifications, and no regulatory guidance or case law currently resolves the question.

\subsubsection{Technical and Operational Challenges}
Implementing Art.~50~II's dual transparency requirements across the \textit{verdict generation}, \textit{validation}, and \textit{release and lifecycle} subsystems presents four interconnected challenges that parallel those in the synthetic data case but arise from structurally different sources. 

First, the fundamental tension between human-readable and machine-readable marking is compounded in fact-checking by the iterative nature of editorial workflows. Text watermarking techniques that could in principle satisfy the machine-readable requirement are highly fragile under the editing, paraphrasing, and summarization practices common in newsrooms \cite{kirchenbauer2023watermark}. A watermark embedded in an AI-generated justification is unlikely to survive the revisions a journalist applies before publication. Unlike the synthetic data case, where a dual-version solution can separate marked and unmarked outputs at distribution time, no equivalent separation point exists in editorial workflows: the human review process intervenes between AI generation and publication, and it is precisely this intervention that destroys the mark. The regulatory requirement for persistent marking directly contradicts the operational reality of human-AI collaboration in journalism. 

Second, the robustness of marking is structurally undermined by RAG-based multi-source attribution. When a justification draws on multiple retrieved sources, weighted by retrieval confidence and filtered by a veracity classifier, no current metadata schema provides a machine-readable format adequate to capture that provenance chain. Established schemas such as Dublin Core or Schema.org lack the semantics for expressing confidence-weighted, multi-step attribution~\cite{kirchenbauer2023watermark}. This is analogous to the watermark fragility problem in synthetic data pipelines, but arises from semantic complexity rather than signal degradation: the provenance information exists but cannot be represented in any interoperable, machine-readable form. The real-time demands of breaking-news scenarios compound this further: cryptographic marking schemes that could support fine-grained provenance are computationally expensive and create latency trade-offs incompatible with live editorial workflows. 

Third, the absence of interoperable standards produces the same fragmented implementation landscape identified in the synthetic data case, but with an additional cross-modal dimension. No unified marking approach currently exists for fact-checking outputs that combine text, images, and video, despite Art.~50~II's explicit multimodal scope. Existing provenance frameworks, including C2PA~\cite{c2pa2024}, provide insufficient support for deeply interleaved, text-centric outputs in which human and AI contributions cannot be cleanly separated. A marking solution developed for text-based verdicts will not extend to video fact-checks without significant additional engineering, and no cross-modal standard currently fills this gap. 

Fourth, the scope of Art.~50~II is ambiguous with respect to explanation quality in ways that have no direct parallel in the synthetic data case. Art.~50~V requires that disclosures be provided in a clear and distinguishable manner conforming to applicable accessibility requirements, but provides no guidance on calibrating explanation depth or format to different user groups. Post-hoc explainability methods such as LIME or SHAP generate technically accurate feature attributions but are inaccessible to non-technical users including most journalists and content moderators~\cite{schmitt2024role}. Natural-language explanations generated by LLMs are more accessible but introduce explanation-induced overreliance: fluent but incorrect rationales increase user trust without improving decision quality~\cite{bansal2021does, schmitt2024role}. The result is a one-size-fits-all transparency requirement that is likely to under-serve both technical and non-technical user groups simultaneously, and whose interaction with the hallucination and non-determinism properties of LLMs~\cite{ji2023survey} means that legally verifiable quality guarantees cannot be provided under current technology. 

Finally, human-AI collaboration introduces additional operational risks. Newsroom environments characterized by time pressure are particularly susceptible to automation bias, potentially eroding independent verification practices and diffusing accountability when AI-assisted fact-checks prove incorrect~\cite{10.1145/3686962}. Although Art.~14's human oversight requirements and Art.~50's transparency obligations are conceptually complementary, concrete guidance on their coordinated implementation in fact-checking contexts remains largely absent. 
Taken together, these four challenges demonstrate that Art.~50~II compliance for automated fact-checking cannot be achieved through marking solutions appended to existing pipelines. The provenance chain that machine-readable marking requires is broken by the editorial workflows that human oversight demands, the semantic complexity of RAG-based attribution exceeds what current standards can represent, and the explanation requirements of Art.~50~V cannot be met without user-group-specific transparency designs that the regulation does not specify. As in the synthetic data case, compliance must be treated as an architectural requirement integrated from the outset, not a post-hoc addition. The next section synthesizes the structural gaps common to both cases and identifies the research directions needed to close them.

\section{Structural Gaps and Research Agenda for Art.~50~II Compliance}\label{sec:Pathways}
The use case analyses show three structural gaps that any compliance pathway must address: the absence of cross-platform marking formats for interleaved human-AI outputs~(RQ1), the misalignment between the regulation's reliability criterion and probabilistic model behavior~(RQ2), and missing guidance for adapting transparency to heterogeneous user expertise~(RQ2). 

\textbf{Pathway 1: Interoperable provenance standards.} No current standard can represent provenance in outputs where human and AI contributions are interleaved. In synthetic data pipelines, the dual-version API approach partially addresses distribution-time separation but lacks cross-platform verification. In fact-checking pipelines, RAG-based attribution cannot be represented in schemas such as Dublin Core or Schema.org, and editorial workflows destroy provenance chains at the point of human review. Future work should extend frameworks such as C2PA~\cite{c2pa2024} to interleaved text-centric outputs and establish robustness benchmarks for marking methods under realistic processing conditions. 

\textbf{Pathway 2: Operational feasibility criteria.} The undefined quality criteria in Art.~50~II leave providers without a clear compliance pathway. In synthetic data contexts, it is unclear whether training-time watermark removal violates the reliability criterion. In fact-checking contexts, no guidance exists on whether marks destroyed by editorial revision satisfy the regulation's persistence requirement. The Commission's Code of Practice~\cite{ec2025codeofpractice} represents an initial step but does not resolve these ambiguities. Future work should develop measurable criteria differentiated by data type, deployment context, and recipient, including the unaddressed distinction between synthetic data for human recipients and for model training pipelines. 

\textbf{Pathway 3: User-group-specific transparency designs.} Art.~50~V requires clear and distinguishable disclosures but provides no guidance on calibrating explanations to different user groups. Post-hoc methods such as LIME or SHAP are inaccessible to non-technical users, while natural-language LLM explanations introduce explanation-induced overreliance~\cite{bansal2021does, schmitt2024role}. Future work should develop and validate disclosure frameworks that adjust explanation depth and format to the recipient's role and expertise, including in time-pressured editorial environments where automation bias poses particular risks~\cite{10.1145/3686962}. These pathways are interdependent: interoperable standards are a prerequisite for machine-readable compliance, operational criteria are necessary to evaluate conformity, and user-centered designs determine whether the human-understandable dimension of dual transparency is achieved in practice. Together they define the minimum research agenda needed before Art.~50~II enters into force in August~2026.

\section{Conclusion}
\label{sec:conclu}
Using synthetic data generation and automated fact-checking as diagnostic use cases, this paper shows that Art.~50~II compliance cannot be reduced to post-hoc labeling. In synthetic data pipelines, persistent marking conflicts with model training integrity; in fact-checking workflows, human editorial intervention destroys the provenance chains that machine-readable marking requires, and no equivalent architectural solution exists. 

For RQ1, some Art.~50~II requirements are achievable where outputs can be accompanied by structured provenance artefacts that support cross-system verification. Systematic limitations arise wherever the regulation presupposes mark persistence and robustness under realistic transformations: precisely the conditions both use cases expose. For RQ2, legal understandability diverges from technical explainability in that explainability methods target model behavior rather than informed use at the point of access. Legal reliability diverges from quality assurance in that probabilistic, non-deterministic LLM outputs cannot provide the verifiable consistency guarantees the regulation implies. 

The compliance consequence is concrete: treating fluent model-generated rationales as evidence of reliability risks explanation-induced overreliance and weakens the regulatory objective of Art.~50~II. These findings support the paper's central argument: dual transparency must be treated as an architectural design requirement integrated across the full AI lifecycle, not a labeling add-on. Closing the three structural gaps identified in Section~\ref{sec:Pathways} requires coordinated action across legal semantics, AI engineering, and human-centered design before Art.~50~II enters into force in August~2026.

\section*{Acknowledgements}
This research is funded by the Federal Ministry of Research, Technology, and Space (BMFTR) in the scope of the research projects news-polygraph (reference: 03RU2U151C), VeraXtract (reference: 16IS24066), and BIFOLD project FakeXplain. Moreover, this paper includes work carried out within the AgrifoodTEF-DE project. AgrifoodTEF-DE (reference: 28DZI04A23) is supported by funds of the Federal Ministry of Agriculture, Food and Regional Identity (BMLEH) based on a decision of the Parliament of the Federal Republic of Germany via the Federal Office for Agriculture and Food (BLE) under the research and innovation program 'Climate Protection in Agriculture'.

\noindent \textbf{Disclosure:} Grammarly, an AI writing assistance tool was used to improve the orthography and grammar of several paragraphs of text.

\section{Bibliographical References}\label{sec:reference}
\vspace{-0.8cm}
\bibliographystyle{lrec2026-natbib}
\bibliography{cleanBib}

\end{document}